\documentclass[wcp]{jmlr}

\usepackage{longtable}% for long tables

\usepackage{booktabs}
\usepackage{mathrsfs}

\jmlrvolume{60}
\jmlryear{2016}
\jmlrworkshop{ACML 2016}

\title[Reinforcement-based Simultaneous Algorithm and its Hyperparameters Selection]{Reinforcement-based Simultaneous Algorithm and its Hyperparameters Selection}

  \author{\Name{Valeria Efimova} \Email{efimova@rain.ifmo.ru}\\
   \Name{Andrey Filchenkov} \Email{afilchenkov@corp.ifmo.ru}\\
   \Name{Anatoly Shalyto} \Email{shalyto@mail.ifmo.ru}\\
   \addr ITMO University, St. Petersburg, Kronverksky Pr. 49}

\DeclareMathOperator*{\argmin}{argmin}
\DeclareMathOperator*{\argmax}{argmax}

\begin{document}

\maketitle

\begin{abstract}
Many algorithms for data analysis exist, especially for classification problems. To solve a data analysis problem, a proper algorithm should be chosen, and also its hyperparameters should be selected. In this paper, we present a new method for the simultaneous selection of an algorithm and its hyperparameters. In order to do so, we reduced this problem to the multi-armed bandit problem. We consider an algorithm as an arm and algorithm hyperparameters search during a fixed time as the corresponding arm play. We also suggest a problem-specific reward function. We performed the experiments on 10 real datasets and compare the suggested method with the existing one implemented in Auto-WEKA. The results show that our method is significantly better in most of the cases and never worse than the Auto-WEKA.
\end{abstract}

\section{Introduction}

%AAF read 15.09
The goal of supervised learning is to find a data model for a given dataset that allows to make the most accurate predictions. To build such model, lots of \emph{learning algorithms} exist, especially in classification. These algorithms show various performances on different tasks. It prevents usage of a single universal algorithm to build a data model for all existing datasets. The performance of most of these algorithms depends on \emph{hyperparameters}, the selection of which dramatically affects the performance of the algorithms. 

%AAF read 14.09
Automated simultaneous selection of a learning algorithm and its hyperparameters is a sophisticated problem. Usually, this problem is divided into two subproblems that are solved independently: algorithm selection and hyperparameter optimization. The first is to select an algorithm from a set of algorithms (algorithm portfolio). The second is to find the best hyperparameters for preselected algorithm.

%AAF read 14.09
The first subproblem is typically solved by testing each of the algorithms with prechosen hyperparameters in the portfolio by many practitioners. Other methods are also in use, such as selecting algorithms randomly, by heuristics or using \emph{k-fold cross-validation}~\citep{kfoldcv}. But the last method requires running and then comparing all the algorithms. The other methods are not universally applicable. However, this subproblem has been in the scope of research interest for decades. Decision rules were used in several decades old papers on algorithm selection from a portfolio~\citep{aha}. As an example, such rules are created to choose from 8 algorithms in~\citep{alism}.

%AAF read 15.09
Nowadays, more effective approaches exists such as meta learning~\citep{metalearn,abdulrahman2015algorithm}. This approach is to reduce the algorithm selection problem to a supervised learning problem. It requires a training set of datasets $D$. For all $d \in D,$ meta-feature vector is evaluated. Meta-features are useful characteristics of datasets, such as number of categorical or numerical features of an object $x \in d$, size of $d$ and many others~\citep{afil2015,castiello2005meta}. After that, all the algorithms are run on all the datasets $d \in D$. Thus class labels are formed based on empirical risk evaluation. Then a meta-classifier is learnt on the prepared data with datasets as objects and best algorithms as labels. It is worth to note that it is better to solve this problem as the learning to rank problem~\citep{brazdil2003ranking,cashmeta2}.

%AAF read 15.09
The second subproblem is a hyperparameter optimization that is to find hyperparameter vector for a learning algorithm that leads to the best performance of this algorithm for a given dataset. For example, hyperparameters of the Support Vector Machine (SVM) include kernel function and its hyperparameters; for a neural net, they include the number of hidden layers and the number of neurons in each of them. In practice, algorithms hyperparameters are usually chosen manually~\citep{hutman}. Moreover, sometimes the selection problem can be reduced to a simple optimization problem (primarily for statistical and regression algorithms), as, for instance, in~\citep{strijov}. However, this method is not universally applicable. Since hyperparameter optimization of classification algorithms is often applied manually, it requires a lot of time and do not lead to acceptable performance. There are several algorithms to solve the second subproblem automatically: Grid Search~\citep{random}, Random Search~\citep{grid}, Stochastic Gradient Descent~\citep{sgd}, Tree-structured Parzen estimator~\citep{tpe}, and the Bayesian Optimization including Sequential Model-Based Optimization (SMBO)~\citep{bayes1}. In~\citep{smac}, Sequential model-based algorithm configuration (SMAC) is introduced. It is based on SMBO algorithm. Another idea is implemented in predicting the best hyperparameter vector with meta-learning approach~\citep{mantovani2015meta}. Reinforcement-based approach was used in~\citep{jamieson2015non} to operate several optimization threads with different settings.

%AAF read 14.09
Solution for the simultaneous selection of an algorithm and its hyperparameters is important for machine learning applications, but only a few of papers are devoted to this search. Moreover, these papers consider only a special case. 

%AAF read 15.09
One of the possible solutions is to build a huge set of algorithms with prechosen hyperparameters and select from it. This solution was implemented in~\citep{cashmeta1}, in which a set of about 300 algorithms with chosen hyperparameters was used. However, such pure algorithm selection approach cannot provide any insurance of these algorithms quality for a new problem. This set may simply not include a hyperparameter vector for one of the presented learning algorithms with the best performance.

%AAF read 15.09
Another possible solution is sequential optimization of hyperparameters for every learning algorithm in portfolio and selection the best of them. This solution is implemented in the Auto-WEKA library~\citep{autowekap}, it allows to choose one of the 27 base learning algorithms, 10 meta-algorithms and 2 ensemble algorithms and optimize its hyperparameters with SMAC method simultaneously and automatically. This method is described in detail in~\citep{autowekap}. It is clear that if we use the method, then it takes enormous time and may be referred to as exhaustive search (while, in fact, it is not due to the infinity of hyperparameter spaces).

%AAF read 15.09
The goal of this work is to suggest a method for simultaneous learning algorithm and its parameters selection being faster than the exhaustive search without affecting found solution quality. In order to do so, we use multi-armed bandit-based approach.

%AAF read 15.09
The remainder of this paper is organized as follows. In Section~\ref{sec:problem}, we describe in details the learning algorithm and its hyperparameter selection problem and its two subproblems. The suggested method, based on multi-armed bandit problem, is presented in Section~\ref{sec:suggest}. In Section~\ref{sec:experiment}, experiment results are presented and discussed. Section~\ref{sec:conclusion} concludes.

This paper extends a paper accepted to International Conference on Intelligent Data Processing: Theory and Applications 2016.

\section{Problem Statement}
\label{sec:problem}

%AAF read 15.09
Let $\Lambda$ be a hyperparameter space related to a learning algorithm $A$. We will denote the algorithm with prechosen hyperparameter vector $\lambda \in \Lambda$ as $A_{\lambda}$.

%AAF read 14.09
Here is the formal description of the algorithm selection problem. We are given a set of algorithms with chosen hyperparameters $\mathcal{A} = \{A^1_{\lambda_1}, \dots A^m_{\lambda_m}\}$ and learning dataset $D = \{d_1, \dots d_n\}$, where $d_i = (x_i, y_i)$ is a pair consisting of an object and its label. We should choose a parametrized algorithm  $A^*_{\lambda^*}$ that is the most effective with respect to a quality measure $Q$. Algorithm efficiency is appraised by the use of dataset partition into learning and test sets with the further empirical risk estimation on the test set. 

%AAF read 15.09
\[
Q(A_{\lambda}, x) = \frac {1} {|D|} \sum_{x \in D} L(A_{\lambda}, x),
\]
where $L(A_{\lambda}, x)$ is a loss function on object $x,$ which is usually $L(A_{\lambda}, x) = [A_{\lambda}(x) \neq y(x)]$
for classification problems.

%AAF read 15.09
The algorithm selection problem thus is stated as the empirical risk minimization problem:
\[
A^*_{\lambda_*} \in \argmin_{A^j_{\lambda_j} \in \mathcal{A}} Q(A^j_{\lambda_j}, D).
\]
Hyperparameter optimization is the process of selecting hyperparameters $\lambda^* \in \Lambda$ of a learning algorithm $A$ to optimize its performance. Therefore, we can write:
\[
\lambda^* \in \argmin_{\lambda \in \Lambda} Q(A_{\lambda}, D).
\]
In this paper, we consider the simultaneous algorithm selection and hyperparameters optimization. We are given learning algorithm set $\mathscr{A}=\{A^1, \dots , A^k\}.$ Each learning algorithm $A^i$ is associated with hyperparameter space $\Lambda^i$. The goal is to find algorithm $A^*_{\lambda^*}$ minimizing the empirical risk:
\[
A^*_{\lambda^*} \in \argmin_{A^j \in \mathscr{A}, \lambda \in \Lambda^j} Q(A_{\lambda}^j, D).
\]

%AAF read 15.09
We assume that hyperparameter optimization is performed during the sequential hyperparameter optimization process. Let us give formal description. \emph{Sequential hyperparameter optimization process} for a learning algorithm $A^i$: 
\[
\pi_i(t, A^i, \{\lambda^i_j\}^k_{j=0}) \rightarrow \lambda^i_{k+1} \in \Lambda^i.
\]
It is a hyperparameter optimization method run on the learning algorithm $A^i$ with time budget $t$, also it stores best found hyperparameter vectors within previous $k$ iterations $\{\lambda_j\}^k_{j=0}$. 

All of the hyperparameter optimization methods listed in the introduction can be described as a sequential hyperparameter optimization process, for instance, Grid Search or any of SMBO algorithm family including SMAC method, which is used in this paper. 
 
%AAF read 15.09
Suppose that a sequential hyperparameter optimization process $\pi_i$ is associated with each learning algorithm $A_i.$ Then the previous problem can be solved by running all these processes. However, a new problem arises, the best algorithm search time minimization problem. In practice, there is a similar problem that is more interesting in practical terms. It is the problem of finding the best algorithm by fixed time. Let us describe it formally.

%AAF read 15.09
Let $T$ be a time budget for the best algorithm $A^*_{\lambda^*}$ searching. We should split $T$ into intervals $T = t_1 + \dots + t_m$ such that if we run process $\pi_i$ with time budget $t_i$ we will get minimal empirical risk.
\[
\min_{j} Q(A_{\lambda_j}^j, D)\xrightarrow
[\left(t_1, \ldots, t_m\right)]{} \min,
\]
where $A^j \in \mathscr{A}, \lambda_j=\pi_j(t_j, A^j, \emptyset)$ and ${t_1+\ldots+t_m=T;\, t_i\ge 0\, \forall i}.$

\section{Suggested method}
\label{sec:suggest}

%AAF read 15.09
In this problem, the key source is a hyperparameter optimization time limit $T$. Let us split it up to $q$ equal small intervals $t$ and call them \emph{time budgets}. Now we can solve time budgets assignment problem. Let’s have a look at our problem in the different way. For each time interval, we should choose a process to be run during this interval before this interval starts.

%AAF read 15.09
The quality that will be reached by an algorithm on a given dataset is a priori unknown. On the one hand, the time spent for searching hyperparameters of not the best learning algorithms is subtracted from the time spent to improve hyperparameters for the best learning algorithm. On the other hand, if the time will be spent only for tuning single algorithm, we may miss better algorithms. Thus, since there is no marginal solution, the problem seems to be to find a tradeoff between exploration (assigning time for tuning hyperparameters of different algorithms) and exploitation (assigning time for tuning hyperparameters of the current best algorithm). This tradeoff detection is the classical problem in reinforcement learning, a special case of which is multi-armed bandit problem~\citep{reinfl}. We cannot assume that there is a hidden process for state transformation that affects performance of algorithms, thus we may assume that the environment is static.   

% AAF read 15.09
Multi-armed bandit problem is a problem, in which there are $N$ bandit's arms. Playing each of the arms grants a certain reward. This reward is chosen according to an unknown probability distribution, specific to this arm. At each iteration $k,$ an agent chooses an arm $a_i$ and get a reward $r(i, k)$. The agent's goal is to minimize the total loss by time $T$. In this paper, we use the following algorithms solving this problem~\citep{reinfl}:
\begin{enumerate}
\item $\varepsilon$-greedy: on each iteration, average reward $\bar r_{a, t}$ is estimated for each arm $a.$ Then the agent plays the arm with maximal average reward with probability $1 - \varepsilon$, and a random arm with probability $\varepsilon.$ If you play each arm an infinite number of times, then the average reward converges to the real reward with probability $1.$

% AAF read 15.09
\item  UCB1: initially, the agent plays each arm once. On iteration $t,$ it plays arm $a_t$ that:
\[
a_t \in \argmax_{i=1..N}  \overline {r_{i,t}} + \sqrt{\frac {2 \cdot \ln t} {n_i}}, 
\]  
where $\overline {r_{i,t}}$ is an average reward for arm $i$, $n_i$ is the number of times arm $i$ was played. 

% AAF read 15.09
\item Softmax: initially, the agent plays each arm once. On iteration $t,$ it plays arm $a_i$ with probability:
\[
p_{a_i} = \frac{e^{{\bar{r_i} / \tau}}}{\sum_{j=1}^N e^{r_j / \tau}},
\]
where $\tau$ is positive temperature parameter. %When $\tau \rightarrow 0$, Softmax is similar to the greedy algorithm.
\end{enumerate}

% AAF read 15.09
In this paper, we associate arms with sequential hyperparameters optimization processes~$\{\pi_i(t, A^i, \{\lambda_k\}^q_{k=0}) \rightarrow \lambda^i_{q+1} \in \Lambda^i\}_{i=0}^m$ for learning algorithms $\mathscr{A} = \{A^1, \dots, A^m\}$. After playing arm $i = a_k$ at iteration $k,$ we assign time budget $t$ to a process $\pi_{a_k}$ to optimize hyperparameters. When time budget runs out, we receive hyperparameter vector $\lambda^i_k$. Finally, when selected process stops, we evaluate the result using empirical risk estimate for process $\pi_i$ at iteration $k,$ that is $Q(A^i_{\lambda^i_k}, D)$.

% AAF read 15.09
The algorithm we name MASSAH (\textbf{M}ulti-\textbf{a}rmed \textbf{s}imultanous \textbf{s}election of \textbf{a}lgorithm and its \textbf{h}yperparameters) is presented listing~\ref{alg:massah}. There, \textsc{MABSolver} is implementing a multi-armed bandit problem solution, \textsc{getConfig}$(i)$ is a function that returns $A^i_{\lambda_q},$ which is the best found configuration by $q$ iterations to algorithm $A^i$.

\begin{algorithm}[t]
  \caption{MASSAH}\label{alg:massah}
    \KwData{$D$ is the given dataset
    $q$ is the number of iterations,\\
    $t$ is time budget for one iteration,\\
    $\{\pi_i\}_{i=1,\ldots,N}$ are sequential hyperparameter optimization processes.\\}
    \KwResult{$A_{\lambda}$ is algorithm with chosen hyperparameters}
    \BlankLine
    \For{$i=1, \dots, N$}{
        $\lambda_i \gets \pi_i(t, A^i, \emptyset)$\\
        $e_i \gets Q(A^i_{\lambda_i}, D)$  
    }
    $best\_err \gets \min_{i=1,\ldots,N} e_i$\\
    $best\_proc \gets \argmin_{i=1,\ldots,N} e_i$\\
    \For{$j = 1, \ldots, q$}{
        $i \gets$ \textsc{MABSolver}$(\{\pi_i\}_{i=1,\ldots,N})$\\
        $\lambda_i \gets \pi_i(t, A^i, \{\lambda_k\}_{k=1}^j)$\\
        $e_i \gets Q(A^i_{\lambda_i}, D)$\\
        \If {$e_i < best\_err$}{
        	$best\_err \gets e_i$\\
        	$best\_proc \gets i$
        }
    }
    \Return \textsc{getConfig}$(\pi_{best\_proc})$
\end{algorithm}

%AAF read 15.09
The question we need to answer is how to define a reward function. The first (and simplest) way is to define a reward as the difference between current empirical risk and optimal empirical risk found during previous iterations. However, we meet several disadvantages. When the optimization process finds hyperparameters that leads to almost optimal algorithm performance, the reward will be extremely small. Also, the selection of such a reward function does not seem to be a good option for MABs, since probability distribution will depend on the number of iterations.

% AAF read 15.09
In order to find a reward function, such that the corresponding probability distribution will not change during the algorithm performance, we apply a little trick. Instead of defining reward function itself, we will define an average reward function. In order to do so, we use SMAC algorithm features.

%AAF read 15.09
Let us describe SMAC algorithm. At each iteration, a set of current optimal hyperparameter vectors is known for each algorithm. A local search is applied to find hyperparameter vectors which have distinction in one position with an optimal vector and improve algorithm quality. These hyperparameter vectors are added to the set. Moreover, some random hyperparameter vectors are added to the set. Then selected configurations (the algorithms with their hyperparameters) are sorted by \emph{expected improvement} (EI). Some of the best configurations are run after that.

As in SMAC, we use empirical risk expectation at iteration $k$: $E_t(Q(A^i_{\lambda^i_k}, D))$, where $Q(A^i_{\lambda^i_k}, D)$ is empirical risk value reached by process $\pi_i$ on dataset $D$ at iteration $k$.

%AAF read 15.09
Note that process $\pi_i$ optimizes hyperparameters for empirical risk minimization, but a multi-armed bandit problem is maximization problem. Therefore, we define an average reward function as:  
\[
\bar r_{i, (k)} = \frac {Q_{max} - E_{(k)}(Q(A^i_{\lambda^i_k}, D))} {Q_{max}},
\]
where $Q_{max}$ is the maximal empirical risk that was achieved on a given dataset.

\section{Experiments}
\label{sec:experiment}

%AAF read 15.10
Since Auto-WEKA implements the only existing solution, we choose it for comparison. Experiments were performed on 10 different real datasets with a predefined split into training and test data from UCI repository\footnote{\url{http://www.cs.ubc.ca/labs/beta/Projects/autoweka/datasets/}}. These datasets characteristics are presented in Table~\ref{tab:dat}.

%AAF read 15.10
\begin{table}[t]
\caption{Datasets description.}\label{tab:dat}
\begin{center}
\begin{tabular}{|l|c|c|c|c|c|}
\hline
Dataset & Number of        & Number of    &  Number    & Number of       &  Number of \\  
        & categorical      & numerical    & of classes & objects in      &  objects in  \\
        & features         & features     &            & training set    &  test set  \\
\hline
Dexter & 0 & 20000 & 2 & 420 & 180 \\
\hline
German Credit & 13 & 7 & 2 & 700 & 300 \\
\hline
Dorothea & 0 & 100000 & 2 & 805 & 345 \\
\hline
Yeast & 0 & 8 & 10 & 1039 & 445 \\
\hline
Secom & 0 & 590 & 2 & 1097 & 470 \\
\hline
Semeion & 0 & 256 & 10 & 1116 & 477 \\ 
\hline
Car & 6 & 0 & 4 & 1210 & 518 \\
\hline
KR-vs-KP & 36 & 0 & 2 & 2238 & 958 \\
\hline
Waveform & 0 & 40 & 3 & 3500 & 1500 \\
\hline
Shuttle & 38 & 192 & 2 & 35000 & 15000 \\
\hline
\end{tabular}
\end{center}
\end{table}

%AAF read 15.09
The suggested approach allows to use any hyperparameter optimization method. In order to perform comparison properly, we use SMAC method that is used by Auto-WEKA. We consider 6 well-known classification algorithms: $k$ Nearest Neighbors (4 categorical and 1 numerical hyperparameters), Support Vector Machine (4 and 6), Logistic Regression (0 and 1), Random Forest (2 and 3), Perceptron (5 and 2), and C4.5 Decision Tree (6 and 2). %The number of categorical and numerical hyperparameters of these algorithms are presented in Table~\ref{tab:hyp}.

%AAF read 15.09
%\begin{table}[t]
%\caption{Number of categorical and numerical hyperparameters for learning algorithms.}\label{tab:hyp}
%\begin{center}
%\begin{tabular}{|l|c|c|}
%\hline
%Algorithm & Categorical & Numerical  \\  
%\hline
%kNN & 4 & 1 \\
%\hline
%SVM & 4 & 6 \\
%\hline
%LR & 0 & 1 \\
%\hline
%RF & 2 & 3 \\
%\hline
%Perceptron & 5 & 2 \\
%\hline
%C4.5 & 6 & 2 \\ 
%\hline
%\end{tabular}
%\end{center}
%\end{table} 

%AAF read 15.09
As we previously stated, we are given time $T$ to find the solution of the main problem. The suggested method requires splitting $T$ into small equal intervals $t$. We give the small interval to a selected process $\pi_i$ at each iteration.
%
%AAF read 15.09
We compare the method performance for different time budget $t$ values to find the optimal value. We consider time budgets from 10 to 60 seconds with 3 second step. After that we run the suggested method on 3 datasets Car, German Credits, KRvsKP described above. We use 4 solutions of the multi-armed bandit problem: UCB1, 0.4-greedy, 0.6-greedy, Softmax. We run each configuration 3 times. The results show no regularity, so we assume time budget $t$ as 30 seconds. 

%AAF read 15.09
In the quality comparison, we consider suggested method with the different multi-armed bandit problem solutions: UCB1, 0.4-greedy, 0.6-greedy, Softmax with the na\"ive reward function, and two solutions $UCB1_{E(Q)}$, $Softmax_{E(Q)}$ with the suggested reward function. Time budget on iteration is $t = 30$ seconds, the general time limitation is $T = 3$ hours $= 10800$ seconds. We run each configuration 12 times with random seeds of SMAC algorithm. Auto-WEKA is also limited to 3 hours and selects one of the algorithms we specified above. The experiment results are shown in Table~\ref{tab:res}. 

%AAF read 15.09
\begin{table}[t]
\caption{Comparison of Auto-WEKA and suggested methods for selecting classification algorithm and its hyperparameters for the given dataset. We performed 12 independent runs of each configuration and report the smallest empirical risk $Q$ achieved by Auto-WEKA and the suggested method variations. We highlight with bold entries that are minimal for the given dataset.}\label{tab:res}
\tabcolsep=2pt
\begin{center}
\begin{tabular}{|l|c|c|c|c|c|c|c|}
\hline 
\small{Dataset} & \small{AutoWEKA} & \small{UCB1} & \small{0.4-greedy} & \small{0.6-greedy} & \small{Softmax} & \small{$UCB1_{E(Q)}$} & \small{$Softmax_{E(Q)}$} \\
\hline
\small Car & \small 0.3305 & \bf \small 0.1836 & \bf \small 0.1836 & \bf  \small 0.1836 & \bf \small 0.1836 & \bf \small 0.1836 & \bf \small 0.1836 \\
\hline
\small Yeast & \small 34.13 & \bf \small 29.81 & \bf \small 29.81 & \small 33.65 & \bf \small 29.81 & \bf \small 29.81 & \bf \small 29.81 \\
\hline
\small KR-vs-KP & \small 0.2976 & \bf \small 0.1488 & \bf \small 0.1488 & \bf \small 0.1488 & \bf \small 0.1488 & \bf \small 0.1488 & \bf \small 0.1488 \\
\hline
\small Semeion & \small 4.646 & \bf \small 1.786 & \bf \small 1.786 & \bf \small 1.786 & \bf \small 1.786 & \bf \small 1.786 & \bf \small 1.786 \\
\hline
\small Shuttle & \bf \small 0.00766 & \small 0.0115 & \small 0.0115 & \bf \small 0.00766 & \small 0.0115 & \bf \small 0.0076 & \bf \small 0.0076 \\
\hline
\small Dexter & \small 7.143 & \small 2.38 & \small 2.381 &  \small 2.381 & \small 2.381 & \small 2.381 & \bf \small 0.16 \\
\hline
\small Waveform  & \small 11.28 & \bf \small 8.286 & \bf \small 8.286 & \bf \small 8.286 & \bf \small 8.286 & \bf \small 8.286 & \bf \small 8.286 \\
\hline
\small Secom  & \small 4.545 & \bf \small 3.636 & \small 4.545 & \small 4.545 & \bf \small 3.636 &  \bf \small 3.636 & \bf \small 3.636 \\
\hline
\small Dorothea  & \small 6.676 & \small 4.938 & \small 4.958 & \small 4.938 & \small 4.938 & \small 4.32 & \bf \small 2.469 \\
\hline
\small German Credits  & \small 19.29 & \bf \small 14.29 & \bf \small 14.29 & \small 15.71 & \bf \small 14.29 & \bf \small 14.29 & \bf \small 14.29\\
\hline
\end{tabular}
\end{center}
\end{table}

%AAF read 15.09
The results show that the suggested method is significantly better in most of the cases than Auto-WEKA of the all 10 datasets, because its variations reach the smallest empirical risk. There is no fundamental difference between the results of the suggested method variations. Nevertheless, algorithms $UCB1_{E(Q)}$ и $Softmax_{E(Q)}$, which use the suggested reward function, achieved the smallest empirical risk in most cases. 

%AAF read 15.09
The experiment results show that the suggested approach improves the existing solution of the simultaneous learning algorithm and its hyperparameters selection problem. Moreover, the suggested approach does not impose restrictions on a hyperparameter optimization process, so the search is performed on the entire hyperparameters space for each learning algorithm. It is significant that the suggested method allows to select a learning algorithm with hyperparameters, whose quality is not worse than Auto-WEKA outcome quality. 

%AAF read 15.09
We claim that the suggested method is statistically not worse than Auto-WEKA. To prove this, we carried out Wilcoxon signed-rank test. In experiments, we use 10 datasets which leads to an appropriate number of pairs. Moreover, other Wilcoxon test assumptions are carried. Therefore, we have 6 test checks: comparison of Auto-WEKA and each variation of the suggested method. Since the number of samples is 10, we have meaningful results when untypical results sum $T < T_{0,01} = 5$. We consider a minimization problem, so we test only the best of 12 runs for each dataset. Finally, we have $T = 3$ for the $\varepsilon$-greedy algorithms and $T = 1$ for the others. This proves the statistical significance of the obtained results.

\section{Conclusions}
\label{sec:conclusion}

%AAF read 14.09
In this paper, we suggest and examine a new solution for the actual problem of an algorithm and its hyperparameters simultaneous selection. The proposed approach is based on a multi-armed bandit problem solution. We suggest a new reward function exploiting hyperparameter optimization method properties. The suggested function is better than the na\"ive function in applying a multi-armed bandit problem solutions to solve the main problem. The experiment result shows that the suggested method outperforms the existing method implemented in Auto-WEKA. 

%AAF read 14.09
The suggested method can be improved by applying meta-learning in order to evaluate algorithm quality to preprocess a given dataset before running any algorithm. This evaluation can be used as a prior knowledge of an algorithm reward. Moreover, we can add a context vector to hyperparameters optimization process and use solutions of a contextual multi-armed bandit problem. We can select some datasets by meta-learning and then get the empirical risk estimate and use it as context.

%\subsection{Subsection Title}
%An figure in Fig.~\ref{fig:spiral}
%\begin{figure}[htp]
%\begin{center}
%\includegraphics[width=0.5\textwidth]{./spiral.png}
%\caption{A spiral.}\label{fig:spiral}
%\end{center}
%\end{figure}
 
%An example of citation~\citep{DBLP:conf/acml/2009}.

\acks{Authors would like to thank Vadim Strijov and unknown reviewers for useful comments. The research was supported by the Government of the Russian Federation (grant~074-U01) and the Russian Foundation for Basic Research (project no. 16-37-60115).}

\bibliography{acml16}

\end{document}